\documentclass[journal]{IEEEtran}

\ifCLASSINFOpdf
\else
\fi

\usepackage{cite}
\usepackage{times}
\usepackage{algorithmic}
\usepackage{array}
\usepackage{amssymb}
\usepackage{mdwmath}
\usepackage{mdwtab}
\usepackage{eqparbox}
\usepackage{cite}
\usepackage{url}
\usepackage{multicol,multirow}
\usepackage[cmex10]{amsmath}
\usepackage{makeidx}
\usepackage{diagbox}
\usepackage{rotating,soul}
\usepackage{wrapfig}
\usepackage{booktabs}
\usepackage[usenames,dvipsnames]{color}
\usepackage{ragged2e}
\usepackage{pifont}
\usepackage{overpic}
\makeindex

\usepackage{geometry}
\RequirePackage{silence}
\DeclareGraphicsExtensions{.pdf,.jpg,.png}

\usepackage[pagebackref=false,breaklinks=false,letterpaper=false,colorlinks,bookmarks=false]{hyperref}

\newcommand{\figref}[1]{Fig.~\ref{#1}}
\newcommand{\tabref}[1]{Tab.~\ref{#1}}

\usepackage{pgfplots}
\pgfdeclarelayer{background}
\pgfdeclarelayer{foreground}
\pgfsetlayers{background,main,foreground}

\definecolor{mygray}{gray}{.92}

\def\ie{\emph{i.e.}}
\def\eg{\emph{e.g.}}
\def\etal{{\em et al.}}

\newcommand{\addFig}[1]{}
\newcommand{\addFigs}[1]{}

\graphicspath{{./figs/}}

\newlength\savedwidth

\begin{document}

\title{Dual-scale Enhanced and Cross-generative Consistency Learning for Semi-supervised Medical Image Segmentation}

\author{
Yunqi Gu, 
Tao Zhou,
Yizhe Zhang,
Yi Zhou, 
Kelei He,
Chen Gong,
Huazhu Fu
\thanks{Y. Gu, T. Zhou, Y. Zhang, and C. Gong are with PCA Laboratory, and the School of Computer Science and Engineering, Nanjing University of Science and Technology, Nanjing 210094, China. Y. Zhou is with the School of Computer Science and Engineering, Southeast University, Nanjing 211189, China. K. He is with the Medical School, Nanjing University, Nanjing 210023, China. He is also with the National Institute of Healthcare Data Science at Nanjing University, Nanjing 210023, China. H. Fu is with the Institute of High Performance Computing, A*STAR, Singapore.}
\thanks{Corresponding author: taozhou.ai@gmail.com (\textit{Tao Zhou}).}
}

\markboth{Pattern Recognition}%
{**}

\maketitle

\begin{abstract}

Medical image segmentation plays a crucial role in computer-aided diagnosis. However, existing methods heavily rely on fully supervised training, which requires a large amount of labeled data with time-consuming pixel-wise annotations. Moreover, accurately segmenting lesions poses challenges due to variations in shape, size, and location. To address these issues, we propose a novel Dual-scale Enhanced and Cross-generative consistency learning framework for semi-supervised medical image Segmentation (DEC-Seg). First, we propose a Cross-level Feature Aggregation (CFA) module that integrates cross-level adjacent layers to enhance the feature representation ability across different resolutions. To address scale variation, we present a scale-enhanced consistency constraint, which ensures consistency in the segmentation maps generated from the same input image at different scales. This constraint helps handle variations in lesion sizes and improves the robustness of the model. Furthermore, we propose a cross-generative consistency scheme, in which the original and perturbed images can be reconstructed using cross-segmentation maps. This consistency constraint allows us to mine effective feature representations and boost the segmentation performance. To further exploit the scale information, we propose a Dual-scale Complementary Fusion (DCF) module that integrates features from two scale-specific decoders operating at different scales to help produce more accurate segmentation maps. Extensive experimental results on multiple medical segmentation tasks (polyp, skin lesion, and brain glioma) demonstrate the effectiveness of our DEC-Seg against other state-of-the-art semi-supervised segmentation approaches.
The implementation code will be released at \href{https://github.com/taozh2017/DECSeg}{https://github.com/taozh2017/DECSeg}.

\end{abstract}


\IEEEpeerreviewmaketitle

\section{Introduction}\label{sec:introduction}
Accurate segmentation of organs or lesions from medical images is a crucial task within the realm of medical image processing and analysis. The successful accomplishment of this task holds immense significance for disease diagnosis, organ localization, and lesion segmentation. In recent years, several deep learning models~\cite{zhou2018unetplus,zhou2024uncertainty,yue2022boundary,lei2023shape} have been developed and have shown promising performance in medical image segmentation. However, these models often rely on a fully supervised training strategy, which necessitates a substantial number of pixel-wise annotations. Unfortunately, the annotation process for medical images is both time-consuming and costly due to the need for specialized expertise. To address these challenges, researchers have turned to semi-supervised learning (SSL) as an effective approach to leverage unlabeled data and enhance the performance of segmentation models~\cite{wang2022semi,meng2022dual}. By utilizing SSL techniques, models can exploit the abundance of available unlabeled data in medical imaging datasets, reducing their dependence on fully annotated data. This approach not only alleviates the burden of extensive annotation requirements but also has the potential to improve the model's accuracy in medical image segmentation tasks.

Recently, various semi-supervised medical image segmentation methods~\cite{zhao2022semi,han2022effective,qiu2023federated} based on deep learning have been developed. Among these methods, the consistent constraint is a widely used strategy, by ensuring that the perturbations on unlabeled data should not significantly vary their outputs or predictions. One of the most representative frameworks is the mean teacher (MT)~\cite{tarvainen2017mean}, which designs a perturbation-based consistency loss between the teacher and student models on the unlabeled examples. Inspired by MT, several improved methods focus on designing different perturbation strategies to achieve SSL segmentation. For instance, an uncertainty-aware framework (UA-MT)~\cite{yu2019uncertainty} was proposed to make the student model gradually learn more reliable targets and eliminate unreliable predictions by exploiting the uncertainty information. 
Verma~\etal~\cite{verma2019interpolation} proposed an interpolation consistency training framework, which constrains the prediction at an interpolation of unlabeled points to be consistent with that of the predictions at those points. Chen~\etal~\cite{chen2021semi} developed a cross-pseudo supervision strategy, which initializes two identical networks with different weights and encourages high consistency between the predictions of two different networks with different weights for the same input. Luo~\etal~\cite{luo2021efficient} designed an uncertainty rectified pyramid consistency (URPC) scheme to enable the segmentation model can produce consistent predictions at different scales. Besides, CLCC~\cite{zhao2022cross} was proposed to fuse global and patch-wise information by contrastive learning and consistency constraint. SLCNet~\cite{liu2022semi} was proposed to utilize shape information for the training by feeding pseudo-labels into the network simultaneously. 
Despite the achievement made, semi-supervised medical image segmentation remains challenging due to variations in the shape, size, and location of lesions. Therefore, it is crucial to develop effective and robust semi-supervised learning strategies that leverage a large amount of unlabeled data to boost the medical image segmentation performance.

To this end, we propose a novel \textbf{D}ual-scale \textbf{E}nhanced and \textbf{C}ross-generative consistency learning framework for semi-supervised medical image \textbf{Seg}mentation (\textbf{DEC-Seg}), which fully exploits multi-scale information from the labeled and unlabeled data to enhance segmentation performance. Specifically, we first propose a Cross-level Feature Aggregation (CFA) module to integrate the adjacent layer features in the encoder, and the cross-level aggregated features are further incorporated into the scale-specific decoder. Then, we present a scale-enhanced consistency strategy to encourage the consistency of segmentation maps from the same inputs with different scales. This consistency can guide our segmentation network to learn more powerful features for handling scale variations. Moreover, a Dual-scale Complementary Fusion (DCF) module is further proposed to filter the effective information from different scale features and fuse them to boost segmentation performance. Meanwhile, a cross-generative consistency is presented to constrain the original images and perturbed ones can be reconstructed by cross-segmentation maps, which can effectively harness the knowledge from unlabeled data. Finally, we conduct the comparison experiments on three medical image segmentation tasks, and the results demonstrate the superiority of the proposed model over other state-of-the-art semi-supervised segmentation methods. 

The main contributions of this paper are listed as follows:

\begin{itemize}
\item We propose a novel semi-supervised medical image segmentation framework, which leverages scale-enhanced consistency and cross-generative consistency to exploit the relationships between labeled and unlabeled data for boosting the segmentation model. 

\item A cross-level feature aggregation module is proposed to fuse the cross-level features, which can enhance the representation ability of features within different resolutions.

\item We present a scale-enhanced consistency constraint to reduce the discrepancy between the predicted segmentation maps from different scale inputs. Besides, a cross-generative consistency is designed to constrain that the original and perturbed images can be reconstructed using cross-segmentation maps, which aims to mine effective feature representations and boost the segmentation performance.

\item A dual-scale complementary fusion module is proposed to benifit the better prediction generation through the complementation and fusion of different scale information from the two scale-specific decoders.

\end{itemize}

 \begin{figure*}[t!]
	\centering
    \footnotesize
	\begin{overpic}[width=0.9\textwidth]{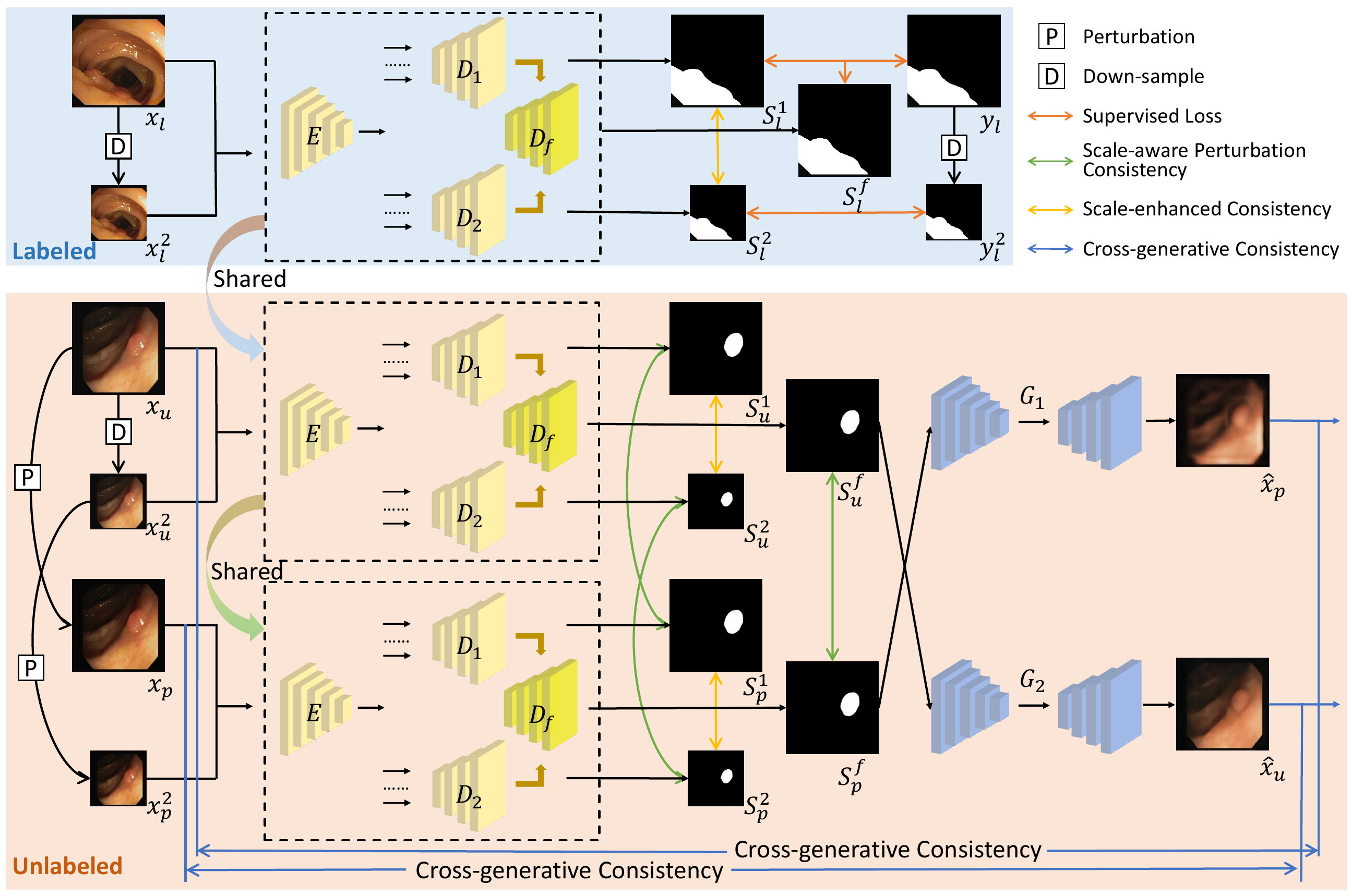}
    \end{overpic}\vspace{-0.25cm}
	\caption{ Architecture of our DEC-Seg framework. Labeled data are fed to the model to achieve the segmentation task under supervised learning. Unlabeled data and the corresponding perturbed versions are passed through the model, which is constrained under scale-enhanced consistency, scale-aware perturbation consistency, and cross-generative consistency.
     See Sec.~\ref{Ourmethod} for details.}
    \label{fig1:Framework}
\end{figure*}

The remainder of this paper is structured as follows. In Section \ref{sec:Related works}, we briefly review some related works to our model, consisting of medical image segmentation, semi-supervised learning, and consistency learning. In Section \ref{Ourmethod}, we give the overview framework and then provide the details of the proposed semi-supervised segmentation model. Then, we present the used datasets, experimental settings, comparison results, and ablation study in Section \ref{Experiments}. Finally, we conclude this paper in Section \ref{Conclusion}.

\section{Related Works}\label{sec:Related works} 

\subsection{Medical Image Segmentation}

Recently, numerous studies have proposed different deep learning architectures and techniques to enhance the accuracy of medical image segmentation. 
The UNet framework, proposed by Ronneberger \etal~\cite{ronneberger2015u}, has significantly advanced the field of medical image segmentation. It incorporates skip connections between the contracting and expanding paths, enabling the network to retain fine-grained spatial information while also capturing high-level contextual features. UNet has demonstrated state-of-the-art performance across various segmentation tasks, including liver and tumor segmentation, cell segmentation, brain tumor segmentation, and polyp segmentation. To further improve the accuracy of UNet-based models, researchers have developed several variants, such as attention UNet~\cite{oktay2018attention}, UNet++~\cite{zhou2018unetplus}, and ResUNet++~\cite{jha2019resunetplus}. For example, Zhou~\etal~\cite{zhou2018unetplus}~presented UNet++ for medical image segmentation which designs a series of nested, dense skip pathways to reduce the semantic gap between the feature maps. Jha~\etal~\cite{jha2019resunetplus}~developed a ResUNet++ architecture, which takes advantage of residual blocks, squeeze and excitation blocks, atrous spatial pyramidal pooling, and attention blocks. Moreover, the cross-level and multi-level feature fusion technique is widely recognized for enhancing segmentation performance by integrating spatial information from low-level features with semantic information from high-level features. For instance, Li \etal~\cite{li2024cifg} proposed a cross-level information processing module that aggregates and processes multi-scale features transmitted by different encoder layers to improve polyp segmentation. Zhou \etal~\cite{zhou2023cross} designed a cross-level feature fusion module to merge adjacent features from various levels, enabling the characterization of cross-level and multi-scale information to address scale variations in polyps. Additionally, CFU-Net~\cite{yin2023cfu} incorporates a multi-level attention module to exploit interactions of contextual information, decision information, and long-range dependencies, thereby enhancing feature propagation within skip connections.

\subsection{Semi-Supervised Learning}

Existing SSL methods can be categorized into self-training, consistency learning, contrastive learning, and adversarial learning. Self-training~\cite{fan2020inf, qiao2018deep} methods use the predictions of the fully-supervised algorithm to act as the pseudo labels of the unlabeled data and retrain the model by mixing with the labeled data. Consistency learning methods~\cite{luo2021efficient,li2021dual,tarvainen2017mean,chen2021semi} enforce agreement between different views to improve the model's performance. By utilizing a consistency loss, these methods ensure that the model's predictions remain consistent across different representations or augmentations of the same input. Contrastive learning methods~\cite{wu2022cross,zhao2022cross,shi2022semi,zhao2023rcps,hu2021semi} obtain better representation learning by contrasting similar features against dissimilar features. Adversarial learning methods~\cite{peiris2021duo,lei2022semi,zhang2017deep} employ a minimax game between a generator and a discriminator to train the model. By generating new and realistic samples, these methods enhance the model's ability to generalize and produce accurate segmentation. Several SSL methods have been developed for the medical image segmentation task. For example, Li~\etal~\cite{li2021dual} proposed a segmentation model with multiple auxiliary decoders and encouraged the consistency of the predictions made by the main decoder and the auxiliary decoders. Besides, contrastive learning~\cite{wu2022cross,chen2024dynamic,tang2024semi} has been widely used in SSL medical image segmentation. Moreover, adversarial learning \cite{peiris2021duo,lei2022semi,chen2022generative} has also emerged to improve the robustness of the model. For example, Peiris~\etal~\cite{peiris2021duo} equipped the network with a critic network to influence the segmentation network to produce the resemble prediction map as ground truth. Lei~\etal~\cite{lei2022semi} proposed double discriminators which are used to learn the prior relationship between labeled and unlabeled data. However, these semi-supervised methods do not deeply explore the exploitability and importance of scale information for semi-supervised learning.

\subsection{Consistency Learning}
Consistency learning as one of the most popular methods in semi-supervised learning aims to force the network to learn potential knowledge from different predictions by applying consistency constraints, which may come from different views of the same input from the same network, or may come from the same input from different networks. Currently, several methods~\cite{tarvainen2017mean,chen2021semi,verma2019interpolation,wu2022mutual,zhong2023semi} based on the mean teacher framework focus on designing different consistent and perturbation strategies to achieve SSL segmentation. For instance, Tarvainen~\etal~\cite{tarvainen2017mean} proposed to generate two different augmented views of the same input and use the output of the teacher model to supervise the student model. Chen~\etal~\cite{chen2021semi} proposed cross-pseudo supervision by enforcing the pseudo supervision between the predictions of two models with different parameter initialization. 
Wu~\etal~\cite{wu2022mutual} proposed three decoders and a different upsampling module in each decoder to amplify the perturbation and impose consistency across the three prediction maps. Zhong~\etal~\cite{zhong2023semi} proposed a multi-attention tri-branch network (MTNet) that each branch uses a different attention mechanism. Among these methods, the consistent constraint is a widely used strategy, aiming to ensure that perturbations applied to unlabeled data do not drastically alter their outputs.

\begin{figure*}[t!]
	\centering
    \footnotesize
	\begin{overpic}[width=2.0\columnwidth]{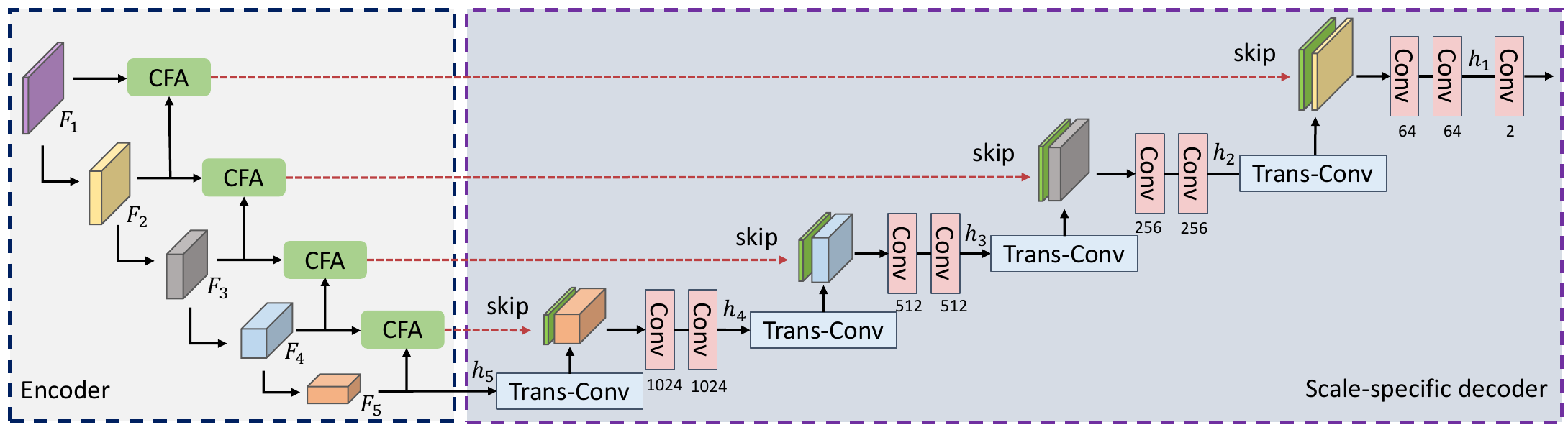}
    \end{overpic}\vspace{-0.05cm}
	\caption{The detailed architectures of the encoder (\ie, $E$ in Fig.~\ref{fig1:Framework}) and scale-specific decoder (\ie, $D_{1}$ and $D_{2}$ in Fig.~\ref{fig1:Framework}) networks. ``Conv" denotes a sequential operation that consists of a $3\times{3}$ convolution, batch normalization, and a \emph{ReLU} activation, and ``Trans-Conv" denotes a transpose convolution.}\vspace{-0.35cm}
    \label{fig2:encoder}
\end{figure*}

\section{Proposed Method}
\label{Ourmethod}

\subsection{Overview}

The proposed DEC-Seg framework, as depicted in \figref{fig1:Framework}, fully leverages multi-scale information from both labeled and unlabeled data for medical image segmentation. Our segmentation network comprises a shared encoder ($E$), two scale-specific decoders ($D_{1}$ processes features solely from the original scale while $D_{2}$ exclusively processes features from the downsampled scale), and a scale-fused decoder ($D_f$ is used to fuse features from both scales). Additionally, two generative networks ($G_1$ and $G_2$) assist in the segmentation process, enhancing the feature mining ability of our segmentation network through a cross-generation strategy. Specifically, the original and downsampled images from the labeled data are fed into the model (\ie, the encoder $E$ and scale-specific decoders $D_{1}$ and $D_{2}$), and then the scale-enhanced consistency and supervised loss are calculated to constrain the network. Then, the unlabeled data and its perturbed versions are inputted into the model. Meanwhile, we conduct scale-enhanced consistency, scale-aware perturbation consistency, and cross-generative consistency for unlabeled data, to fully leverage unlabeled samples to improve the segmentation performance. Moreover, to make use of multi-scale information, we design the scale-fused decoder (\ie, \emph{D$_{f}$}) to generate the final segmentation maps. For convenience, we denote ${X}_l=\left\{x_l^1,x_l^2, \cdots, x_l^{n_l}\right\}$ and ${Y}_l=\left\{y_l^1,y_l^2, \cdots, y_l^{n_l}\right\}$ as the labeled dataset and the corresponding label set, respectively, where $n_l$ is the number of the labeled images. We use ${X}_u=\left\{x_u^1,x_u^2, \cdots, x_u^{n_u}\right\}$ to denote the unlabeled dataset with $n_u$ images, and we typically have $n_l \ll n_u$. Given inputs $X$, they are fed into the encoder to learn five levels of features, namely $\{F_i\}_{i=1}^{5}$.

\subsection{Cross-level Feature Aggregation Module}

\begin{figure}[t!]
	\centering
    \footnotesize
	\begin{overpic}[width=0.95\columnwidth]{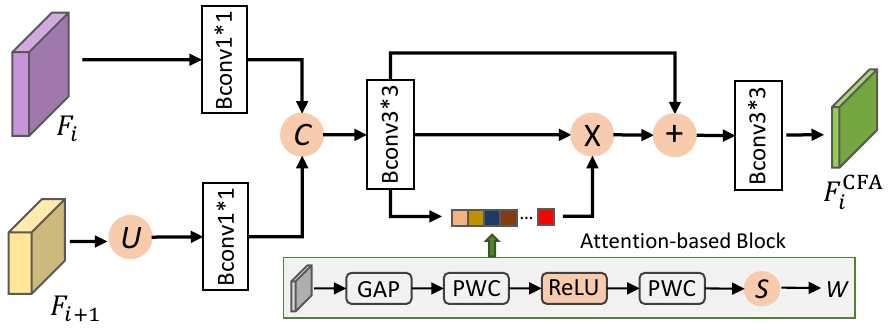}
    \end{overpic}\vspace{-0.35cm}
	\caption{ Illustrations of cross-level feature aggregation module. ``GAP" denotes the global average pooling, ``PWC" denotes the point-wise convolution, and ``$U$" denotes the upsampling using bilinear interpolation.}
    \label{fig3:CFA}
\end{figure}

To fully exploit cross-level information from various receptive fields and provide more effective supplementation from different convolutional layers, we propose the CFA module. This module fuses features from every two adjacent layers, enhancing the understanding of context and relationships within the cross-level features. The aggregated feature is then incorporated into our designed decoder, as depicted in \figref{fig2:encoder}. 
Specifically, as shown in \figref{fig3:CFA}, two adjacent features $F_i$ and $F_{i+1}$ first proceeded by a $1\times 1$ convolution, and then the two features are concatenated (\ie, $F_{cat}$) to obtain $F_{cat}^{'}=\mathcal{B}_{conv3 \times 3}(F_{cat})$, 
where $\mathcal{B}_{conv3{\times}3}(\cdot)$ is a sequential operation that consists of a $3\times{3}$ convolution, batch normalization, and a \emph{ReLU} activation. To learn the cross-level attention-based enhanced feature, we conduct a global average pooling (GAP) on the cascaded feature $F_{cat}^{'}$ and utilize the point-wise convolution (PWC)~\cite{dai2021attentional} to capture channel interactions across different spatial positions. Therefore, we obtain the attention-based weights by 
\begin{equation}
\begin{aligned}
W=\sigma\big(\textup{Conv}_{pwc2}(\zeta(\textup{Conv}_{pwc1}(\mathcal{G}_{ave}(F_{cat}^{'}))))\big),
\end{aligned}
\end{equation}
where the kernel sizes of $\textup{Conv}_{pwc1}$ and $\textup{Conv}_{pwc2}$ denote as $\frac{C}{r}\times C\times 1 \times 1$ and $ C\times \frac{C}{r}\times 1 \times 1$ ($C$ is channel size and $r$ is a reduction ratio), respectively. Besides, $\zeta(\cdot)$ and $\sigma(\cdot)$ indicate \emph{ReLU} and \emph{Sigmoid} activation functions, respectively, and $\mathcal{G}_{ave}$ is a GAP operation. Next, an element-wise multiplication is utilized to enhance $F_{cat}^{'}$ with $W$, and then a residual structure is also adopted to fuse the enhanced feature and the original cascaded feature. Finally, we obtain the aggregated feature by
\begin{equation}
\begin{aligned}
F_i^{\textup{CFA}}=\mathcal{B}_{conv3{\times}3}\big(F_{cat}^{'}\otimes{W} \oplus F_{cat}^{'}\big),
\end{aligned}
\end{equation}
where $\oplus$ and $\otimes$ represent element-wise addition and multiplication, respectively. 
It is worth noting that our CFA module can explore contextual information from the diversity of resolutions to enhance the features' representation ability.

\subsection{Scale-enhanced Consistency}\label{sec:SC}

The scale variation is still a great challenge for medical image segmentation. As discussed in~\cite{Wang2020cvpr}, reducing the gap between the network outputs from different scale images can help the model learn scale-correlated features.
Motivated by this observation, we propose a scale-enhanced consistency scheme to constrain the outputs of different scale images to be closed, in which the different scale features can be refined by each other. 
To increase the perturbation, we use two independent decoders for the single-scale prediction task, which are called scale-special decoders (\ie, $D_1$ and $D_2$). 
Specifically, given unlabeled images $X_{u}$, $X_{u}$ and its downsampled versions $X_{u}^2=Down(X_{u})$ are fed to a weight-shared encoder to extract two sets of multi-level features, and then we can obtain the predicted maps $S_{u}^1$ and $S_{u}^{2}$, respectively. Specifically, the two segmentation maps can be obtained by $S_{u}^1=D_{1}(E(X_{u}))$ and $S_{u}^{2}=D_{2}(E(X_{u}^2))$, where $E(\cdot)$ is the encoder, and $Down(\cdot)$ denotes a $1/2$ downsampling operation. Furthermore, we employ the mutual supervision approach outlined in \cite{chen2021semi} to ensure the consistency of $S^1_u$ and $S^2_u$ with different sizes. Concretely, $\hat{Y}^1_u=\arg\max\limits_{c}(S^1_u)$ and $\hat{Y}^2_u=\arg\max\limits_{c}(S^2_u)$ are computed. Subsequently, $\hat{Y}^1_u$ is downsampled to the size of $S^2_u$ and utilized as a pseudo-label for $S^2_u$, while $\hat{Y}^2_u$ is upsampled to the size of $S^1_u$ and employed as a pseudo-label for $S^1_u$. Therefore, the scale-enhanced consistency loss for unlabeled data is defined by
\begin{equation}
\mathcal{L}_{SC}^u(S^1_u,S^2_u) = \mathcal{L}_{ce} (S^1_u, Up(\hat{Y}^2_u)) + \mathcal{L}_{ce} (S^2_u, Down(\hat{Y}^1_u)),
\end{equation}
where $\mathcal{L}_{ce}$ denotes the widely used cross entropy loss and $Up(\cdot)$ denotes a $2\times$ upsampling operation.
Similarly, for labeled images $X_l$, we can obtain the predicted segmentation maps $S_{l}^1$ and $S_{l}^{2}$ from $X_l$ and its downsampled versions $X_l^2$ by $S_{l}^1=D_{1}(E(X_{l}))$ and $S_{l}^{2}=D_{2}(E(X_{l}^2))$, respectively. 
As a result, $\mathcal{L}_{SC}^l(S^1_l,S^2_l)$ can be obtained in the same way.

\begin{figure}[!t]
	\centering
	\footnotesize
	\begin{overpic}[width=0.95\columnwidth]{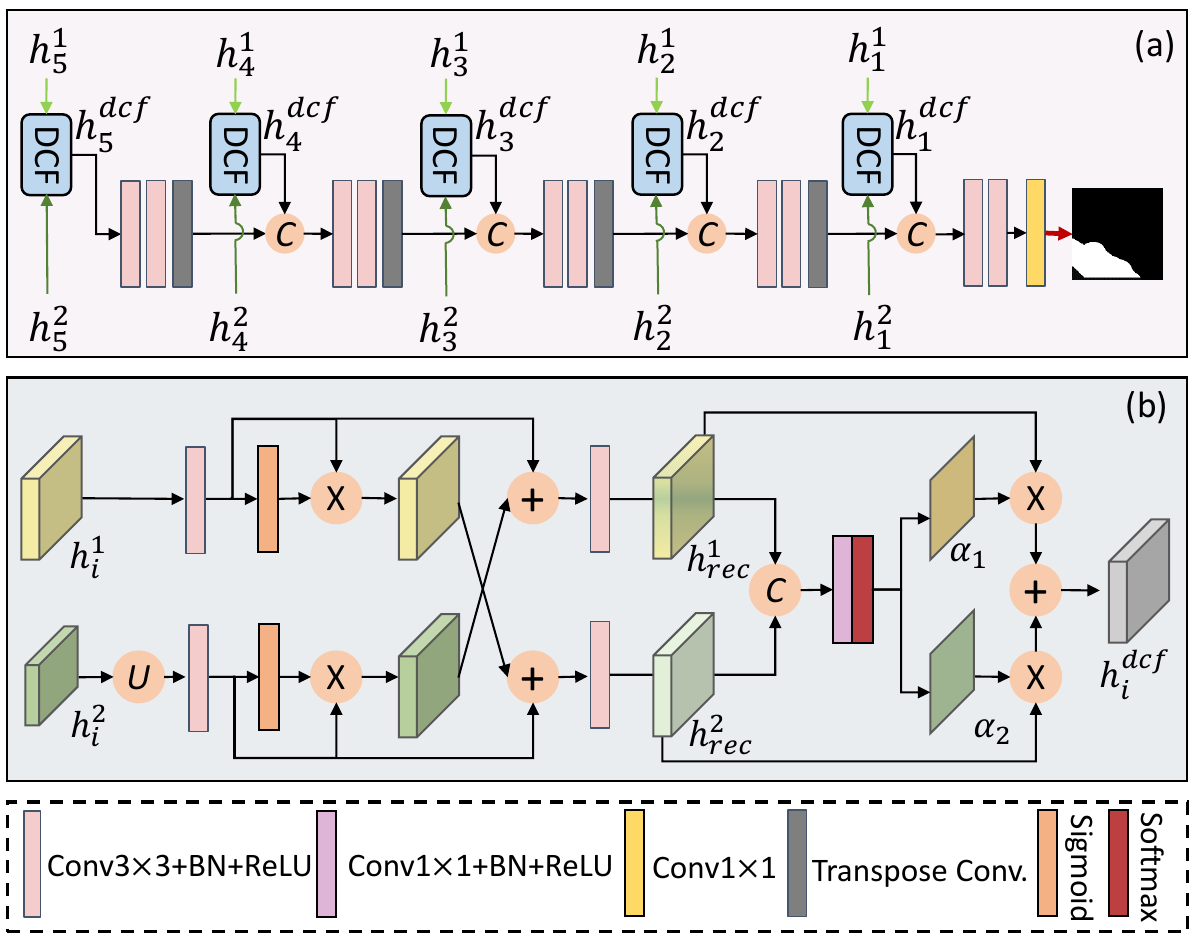}
	\end{overpic}\vspace{-0.25cm}
	\caption{ Illustration of (a) the architecture of the scale-fused decoder (\ie, ${D}_f$ in Fig.~\ref{fig1:Framework}) and (b) the proposed dual-scale complementary fusion module.  ``$U$" denotes the upsampling using a bilinear interpolation.}
	\label{fig4:DCF}
\end{figure}

\subsection{Dual-scale Complementary Fusion Module}

The two independent decoders, as designed in Section \ref{sec:SC}, are specialized for processing information at two distinct scales. Throughout network training, these decoders acquire unique features based on the input image scale, with each emphasizing different aspects. Despite this distinction, the features from both scales pertain to the same unlabeled image, indicating the opportunity to combine and extract features from these two scale features further, potentially resulting in an enhanced segmentation map. Consequently, to fully leverage information from different scales, in comparison to the scale-specific decoder, we also introduce the scale-fused decoder  (\emph{D$_f$} in Fig.~\ref{fig1:Framework}) to produce more reliable predictions as the final segmentation maps. 

As shown in Fig.~\ref{fig1:Framework}, the features from $D_{1}$ and $D_{2}$ are fused and then fed into the scale-fused decoder. The feature flow of the scale-fused decoder is shown in \figref{fig4:DCF}(a), and the structure of the scale-fused decoder is basically the same as that of the scale-specific decoder. To achieve the fuse operation, we present a dual-scale complementary fusion (DCF) module to fuse the features from the two scale-specific decoders. This module replaces the features transmitted from the encoder through the skip operation. The proposed DCF module aims to fuse the multi-scale features and enable the features from the original scale and downsampled scale to complement each other (as shown in \figref{fig4:DCF}(b)). Taking the original scale feature as an example, we first conduct a $3\times3$ convolution on $Up(h_i^{2})$ and Sigmoid activation function to obtain a scale-aware weight $W_2=\sigma(\mathcal{B}_{conv3 \times 3}(Up(h_i^{2})))$. Then, we multiply $W_2$ by $\mathcal{B}_{conv3 \times 3}(Up(h_i^{2}))$, and the resultant feature is the complementary information (\ie, $h_{com}^2$) which is encouraged to learn the information required by the original scale. Finally, we add the complementary feature obtained at the downsampled scale to the original scale and then smooth the feature by a $3\times3$ convolution, thus we can obtain $h_{rec}^1=\mathcal{B}_{conv3 \times 3}(\mathcal{B}_{conv3 \times 3}(h_i^1)+\mathcal{B}_{conv3 \times 3}(Up(h_i^{2}))*W_2)$. Similarly, we can obtain the complementary-enhanced downsampled feature $h_{rec}^2$. Through a cross-scale complementary-enhanced process, we learn more rich multi-scale feature representations. Further, two features $h_{rec}^1$ and $h_{rec}^2$ are concatenated and then proceeded by a $1\times1$ convolution to two channels and a \emph{Softmax} function to obtain two weight maps $\alpha_1$ and $\alpha_2$, where $\alpha_1+\alpha_2=1$. Finally, we obtain the multi-scale fused feature by
\begin{equation}
    h_i^{dcf}=h_{rec}^1 \otimes \alpha_1 + h_{rec}^2 \otimes  \alpha_2.
\end{equation}

Subsequently, the fused features $h_i^{dcf}$ will concatenate with the features from the previous layer and continue to propagate forward. Specifically, $h_5^{dcf}$ is first passed through two convolutional layers and a transpose convolutional layer to have the same resolution with $h_4^{dcf}$. Then, the two features are concatenated and fed into the next two convolutional layers. Repeating the above process, we will obtain the final segmentation maps $S^f$. It is important to highlight that the scale-fused decoder plays a vital role in fully integrating multi-scale information, consequently enhancing the segmentation performance. Consequently, during the inference stage, the segmentation predictions obtained from the scale-fused decoder are taken as the final segmentation results.

\begin{figure*}[!t]
	\centering
	\footnotesize
	\begin{overpic}[width=2.0\columnwidth]{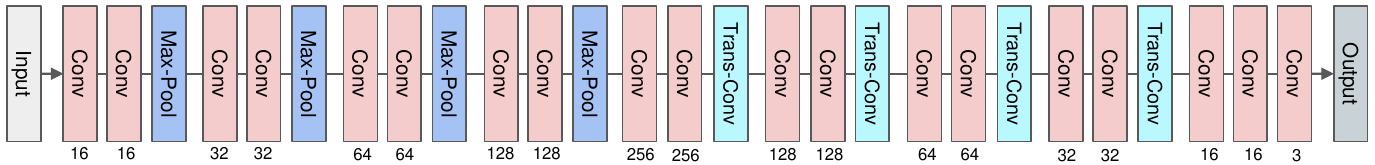}
	\end{overpic}\vspace{-0.15cm}
	\caption{\footnotesize The detailed architecture of the generative network (\ie, $G_{1}$ and $G_{2}$). ``Conv" denotes a sequential operation that consists of a $3\times{3}$ convolution, batch normalization, and a \emph{ReLU} activation, and ``Trans-Conv" denotes a transpose convolution.}
	\label{fig5}
\end{figure*}

\subsection{Scale-aware Perturbation and Cross-generative Consistency}\label{sec:CC}
To constrain the predictions of unlabeled data on the scale-fused decoder, we follow previous works~\cite{li2021dual,wang2022dual,worrall2017harmonic}, and introduce scale-aware perturbation consistency to encourage the outputs between the original version and perturbed version to be closed. Specifically, we impose the perturbation consistency on different scale images, \ie, $S_p^1=D_{1}(E(\mathcal{P}(X_u)))$, $S_p^2=D_{2}(E(\mathcal{P}(X_u^2)))$, and $S_p^f$, where a random color jitter operation is adopted as the perturbation operation (\ie, $\mathcal{P}$). Thus, $\mathcal{L}_{SC}^p(S^1_p,S^2_p)$ is obtained and the scale-aware perturbation consistency scheme can be optimized by minimizing $\mathcal{L}_{SPC}$, which is defined by
\begin{equation} \label{PC}
\mathcal{L}_{SPC} = \mathcal{L}_{mse} (S_u^1, S
_p^1)+\mathcal{L}_{mse}(S_u^2,S_p^2) + \mathcal{L}_{mse} (S_u^f, S_p^f).
\end{equation}

Further, to utilize the rich information contained in different feature maps, we propose a perturbation-based cross-generative consistency constraint to enhance the feature mining ability of our segmentation network with information reconstruction and to correct the cognitive bias on a single input version with cross-generation. Therefore, we design two generative networks $G_1$ and $G_2$ (The detailed structure is shown in Fig.~\ref{fig5}) for image reconstruction. 
For the generative network to accept richer information, produce better-reconstructed maps, and deliver more valuable feedback to the segmentation network, the logit predictions $z_u^f$ and $z_p^f$ obtained from the scale-fused decoder are fed into $G_2$ and $G_1$, respectively, and then we can obtain two reconstructed images $\hat{X}_p=G_1(z_p^f)$ and $\hat{X}_u=G_2(z_u^f)$. Finally, we form a cross-generative consistency loss ($\mathcal{L}_{CC}$), which is given by
\begin{equation} \label{CC}
\mathcal{L}_{CC} = \mathcal{L}_{mse} (\hat{X}_p, X_u)+\mathcal{L}_{mse}(\hat{X}_u,X_p).
\end{equation}

It is worth noting that $\mathcal{L}_{CC}$ can enforce $\hat{X}_p$ and $X_u$ to be consistent and $\hat{X}_u$ and $X_p$ to be closed. Using the input image as a learning target provides unlabeled data with an authentic constraint containing structural and semantic information. This empowers the segmentation network to prioritize the structural and semantic details that are pertinent to category information during the generation of segmentation maps. The cross-generation process compels the network to investigate inconsistent predictions caused by color variations, thereby rectifying the network's cognitive biases toward a single input and ensuring the accuracy of predictions for both the original input and the perturbed input from different perspectives. Consequently, such cross-generation can further enhance the robustness of our model, enabling it to learn powerful features and effectively leverage the knowledge from unlabeled data to enhance the proposed segmentation model.

\subsection{Overall Loss Function}

In this study, the supervised loss for labeled data is the sum of the loss on the prediction maps of the designed three decoders and its corresponding ground truth, and it is formulated by 
\begin{equation}\label{equ:4}
\begin{split}
\mathcal{L}_S=\mathcal{L}_{sup}(S_{l}^1, Y_l) + \mathcal{L}_{sup}(S_{l}^{2}, Down(Y_l)) + \mathcal{L}_{sup}(S_{l}^{f}, Y_l),
\end{split}
\end{equation}
where $\mathcal{L}_{sup} = (\mathcal{L}_{CE} + \mathcal{L}_{Dice})/2$, $\mathcal{L}_{CE}$ and $\mathcal{L}_{Dice}$ denote the cross-entropy (CE) loss and Dice loss, respectively. 
Finally, the total loss can be given as follows:
\begin{equation} 
\mathcal{L}_{total} = \mathcal{L}_S + \mathcal{L}_{SPC}+
\mathcal{L}_{SC} + \mathcal{L}_{CC},
\end{equation} 
where $\mathcal{L}_{SC}=\mathcal{L}_{SC}^l+\mathcal{L}_{SC}^u+\mathcal{L}_{SC}^p$ is the sum of scale-enhanced consistency loss in labeled images, unlabeled images, and perturbed images.

\section{Experiments and Results}\label{Experiments}

\subsection{Datasets}
\textbf{Colonoscopy Datasets}: We carry out the comparison experiments on four public datasets, \ie, CVC-ColonDB~\cite{tajbakhsh2015automated}, Kvasir~\cite{jha2020kvasir}, CVC-ClinicDB~\cite{bernal2015wm}, and ETIS~\cite{silva2014toward}. Following the setting in~\cite{fan2020pranet}, $1,450$ images are randomly selected from the two datasets (\ie, CVC-ClinicDB and Kvasir) for the training set, and the remaining images from the two datasets and the other two datasets (\ie, ETIS and CVC-ColonDB) to form the testing set.
Besides, $10\%$ ($145$ images) or $30\%$ ($435$ images) from the training set are used as the labeled data, while the remaining images are adopted as the unlabeled data. 

\textbf{Brain Tumor Dataset}: This dataset contains brain MR images together with manual FLAIR abnormality segmentation masks and it comes from 110 patients included in The Cancer Genome Atlas (TCGA) lower-grade glioma~\cite{buda2019association, mazurowski2017radiogenomics}. In this study, we have $1,090$ images for the training set and 283 ones for the testing set. In addition, $10\%$ and $30\%$ images from the training set are adopted as the labeled data, respectively, while the remaining training images are taken as unlabeled data.

\textbf{Skin Lesion Dataset}: The dataset comes from ISIC-2018 challenge~\cite{codella2019skin} and is used for skin lesion segmentation. It includes $2,594$ dermoscopic images for training and $1,000$ images for testing. In the same way, $10\%$ or $30\%$ images from the training set are used as the labeled data, while the remaining images are formed as the unlabeled set.

\begin{table*}[t!]
  \centering
  \small
  \renewcommand{\arraystretch}{1.1}
  \setlength\tabcolsep{5.0pt}
  \caption{Quantitative results on CVC-ClinicDB and Kvasir datasets with $10\%$ or $30\%$ labeled data.
  }\label{tab1:seen}\vspace{-0.25cm}
  \begin{tabular}{c|r||ccccc|ccccc}
  \hline
  \multicolumn{2}{c||}{\multirow{2}*{\textbf{Methods}}} & \multicolumn{5}{c|}{10\% labeled} & \multicolumn{5}{c}{30\% labeled}\\
  \cline{3-12}
  \multicolumn{2}{c||}{} & mDice $\uparrow$ & mIoU $\uparrow$ &  $F_\beta^w$ $\uparrow$ & $S_{\alpha}$ $\uparrow$ & MAE $\downarrow$ & mDice $\uparrow$ & mIoU $\uparrow$  & $F_\beta^w$ $\uparrow$ & $S_{\alpha}$ $\uparrow$ & MAE $\downarrow$\\
    \hline
    
  \multirow{12}{*}{\begin{sideways}CVC-ClinicDB\end{sideways}} 
  & MT~\cite{tarvainen2017mean} & 0.747 & 0.662 & 0.741 & 0.830 & 0.028 & 0.772 & 0.689 & 0.763 & 0.852 & 0.028\\
  & DAN~\cite{zhang2017deep} & 0.755 & 0.674 & 0.733 & 0.842 & 0.027 & 0.775  & 0.696 & 0.772 & 0.848 & 0.027\\
  & UA-MT~\cite{yu2019uncertainty} & 0.749 & 0.676 & 0.742 & 0.839 & 0.027 & 0.790 &  0.715  & 0.787 & 0.853 & 0.025\\
  & URPC~\cite{luo2021efficient} & 0.769 & 0.696 & 0.760 & 0.854 &0.021 & 0.777 & 0.709 & 0.785 & 0.853 & 0.026\\
  & CLCC~\cite{zhao2022cross} & 0.794 & 0.720 & 0.786 & 0.859 & 0.027 & 0.840 & 0.786 & 0.840 & 0.891 & 0.019\\
  & SLC-Net~\cite{liu2022semi} & 0.752 & 0.689 & 0.735 & 0.844 & 0.027 & 0.835 & 0.774 & 0.830 & 0.888 & 0.020\\
  & MC-Net+~\cite{wu2022mutual} & 0.767 & 0.702 & 0.751 & 0.856 & 0.025 & 0.845 & 0.783 & 0.844 & 0.891 & 0.020\\
  & CDMA~\cite{zhong2023semi} & 0.759 & 0.676 & 0.750 & 0.837 & 0.025 & 0.803 & 0.728 & 0.802 & 0.863 & 0.024\\
  & SCP-Net~\cite{zhang2023self} & 0.776 & 0.703 & 0.758 & 0.854 & 0.024 & 0.839 & 0.786 & 0.837 & 0.888 & 0.020\\
  & MCF~\cite{wang2023mcf} & 0.779 & 0.718 & 0.767 & 0.860 & 0.021 & 0.823 & 0.766 & 0.819 & 0.882 & 0.019\\
  & \textbf{DEC-Seg} (ours) & \textbf{0.836} & \textbf{0.774} & \textbf{0.834} & \textbf{0.886} & \textbf{0.019} & \textbf{0.859} & \textbf{0.804} & \textbf{0.860} & \textbf{0.900} & \textbf{0.015}\\
  \hline

  \multirow{12}{*}{\begin{sideways}Kvasir\end{sideways}}
  & MT~\cite{tarvainen2017mean} & 0.814 & 0.722 & 0.797 & 0.841 & 0.050 & 0.814 & 0.732 & 0.790 & 0.852 & 0.052\\
  & DAN~\cite{zhang2017deep} & 0.808 & 0.723 & 0.786 & 0.839 & 0.059 & 0.841  & 0.760 & 0.828 & 0.866 & 0.043\\
  & UA-MT~\cite{yu2019uncertainty} & 0.799 & 0.713 & 0.782 & 0.834 & 0.058 & 0.845 &  0.771  & 0.840 & 0.873 & 0.037\\
  & URPC~\cite{luo2021efficient} & 0.811 & 0.728 & 0.796 & 0.842 & 0.057 & 0.849 & 0.778 & 0.846 & 0.874 & 0.040\\
  & CLCC~\cite{zhao2022cross} & 0.806 & 0.724 & 0.796 & 0.844 & 0.052 & 0.864 & 0.804 & 0.853 & 0.892 & 0.034\\
  & SLC-Net~\cite{liu2022semi} & 0.840 & 0.773 & 0.830 & 0.868 & 0.042 & 0.867 & 0.805 & 0.856 & 0.890 & 0.034\\
  & MC-Net+~\cite{wu2022mutual} & 0.817 & 0.735 & 0.807 & 848 & 0.051 & 0.831 & 0.765 & 0.816 & 0.862 & 0.044\\
  & CDMA~\cite{zhong2023semi} & 0.786 & 0.695 & 0.769 & 0.826 & 0.056 & 0.839 & 0.758 & 0.826 & 0.862 & 0.044\\
  & SCP-Net~\cite{zhang2023self} & 0.810 & 0.723 & 0.790 & 0.840 & 0.056 & 0.862 & 0.789 & 0.847 & 0.878 & 0.037\\
  & MCF~\cite{wang2023mcf} & 0.822 & 0.751 & 0.816 & 0.852 & 0.051 & 0.856 & 0.794 & 0.847 & 0.882 & 0.039\\
  & \textbf{DEC-Seg} (ours) & \textbf{0.859} & \textbf{0.787} & \textbf{0.853} & \textbf{0.877} & \textbf{0.044} & \textbf{0.893} & \textbf{0.830} & \textbf{0.886} & \textbf{0.902} & \textbf{0.032}\\
  \hline
  \end{tabular}\vspace{-0.15cm}
\end{table*}

\subsection{Implementation Details and Evaluation Metrics}

\textbf{Implementation Details}: The proposed segmentation framework is implemented in PyTorch and trained on one NVIDIA GeForce RTX3090 GPU. The input images from the polyp segmentation dataset and skin lesion segmentation dataset are uniformly rescaled to $352 \times 352$, and the inputs of the brain MRI segmentation dataset are uniformly rescaled to $256 \times 256$.
Our model converges over $10,000$ iterations with a batch size of $6$ including 3 labeled samples and 3 unlabeled samples, and uses an SGD optimizer with an initial learning rate set to $1e$-$2$ and a poly learning rate strategy to update the learning rate. 

\textbf{Evaluation Metrics}: To evaluate the effectiveness, we employ five commonly adopted metrics~\cite{fan2020pranet,jha2019resunetplus}, namely mean Dice (mDice), mean IoU (mIoU), $F_\beta^w$, $S_{\alpha}$, and mean absolute error (MAE).

\begin{table*}[t!]
  \centering
  \small
  \renewcommand{\arraystretch}{1.1}
  \setlength\tabcolsep{5.0pt}
  \caption{Quantitative results on two unseen datasets (CVC-ColonDB and ETIS) with $10\%$ or $30\%$ labeled data.
  }\label{tab2:unseen}\vspace{-0.25cm}
  \begin{tabular}{c|r||ccccc|ccccc}
  \hline
  \multicolumn{2}{c||}{\multirow{2}*{\textbf{Methods}}} & \multicolumn{5}{c|}{10\% labeled} & \multicolumn{5}{c}{30\% labeled}\\
  \cline{3-12}
  \multicolumn{2}{c||}{} & mDice $\uparrow$ & mIoU $\uparrow$ &  $F_\beta^w$ $\uparrow$ & $S_{\alpha}$ $\uparrow$ & MAE $\downarrow$ & mDice $\uparrow$ & mIoU $\uparrow$  & $F_\beta^w$ $\uparrow$ & $S_{\alpha}$ $\uparrow$ & MAE $\downarrow$\\
    \hline

  \multirow{12}{*}{\begin{sideways}CVC-ColonDB\end{sideways}} 
  & MT~\cite{tarvainen2017mean} & 0.589 & 0.497 & 0.581 & 0.743 & 0.048 & 0.600 & 0.510 & 0.597 & 0.749 & 0.045\\
  & DAN~\cite{zhang2017deep} & 0.620 & 0.517 & 0.604 & 0.756 & 0.046 & 0.633 & 0.546 & 0.630 & 0.767 & 0.045\\
  & UA-MT~\cite{yu2019uncertainty} & 0.553 & 0.471 & 0.551 & 0.727 & 0.051 & 0.648 & 0.563 & 0.635 & 0.771 & 0.046\\
  & URPC~\cite{luo2021efficient} & 0.556 & 0.480 & 0.549 & 0.732 & 0.047 & 0.598 & 0.519 & 0.601 & 0.754 & 0.045\\
  & CLCC~\cite{zhao2022cross} & 0.538 & 0.473 & 0.533 & 0.720 & 0.053 & 0.564 & 0.505 & 0.562 & 0.738 & 0.050\\
  & SLC-Net~\cite{liu2022semi} & 0.595 & 0.525 & 0.582 & 0.755 & 0.044 & 0.653 & 0.582 & 0.644 & 0.783 & 0.041\\
  & MC-Net+~\cite{wu2022mutual} & 0.562 & 0.486 & 0.543 & 0.725 & 0.055 & 0.589 & 0.518 & 0.578 & 0.745 & 0.050\\
  & CDMA~\cite{zhong2023semi} & 0.507 & 0.419 & 0.494 & 0.691 & 0.054 & 0.624 & 0.536 & 0.609 & 0.756 & 0.047\\
  & SCP-Net~\cite{zhang2023self} & 0.577 & 0.495 & 0.559 & 0.739 & 0.049 & 0.694 & 0.615 & 0.683 & 0.802 & 0.039\\
  & MCF~\cite{wang2023mcf} & 0.566 & 0.498 & 0.559 & 0.742 & \textbf{0.045} & 0.621 & 0.550 & 0.614 & 0.768 & 0.042\\
  & \textbf{DEC-Seg} (ours) & \textbf{0.648} & \textbf{0.565} & \textbf{0.630} & \textbf{0.771} & 0.046 & \textbf{0.721} & \textbf{0.640} & \textbf{0.709} & \textbf{0.814} & \textbf{0.035}\\
  \hline

  \multirow{12}{*}{\begin{sideways}ETIS\end{sideways}}
  & MT~\cite{tarvainen2017mean} & 0.356 & 0.288 & 0.337 & 0.636 & 0.035 & 0.495 & 0.420 & 0.462 & 0.706 & 0.039\\
  & DAN~\cite{zhang2017deep} & 0.437 & 0.358 & 0.397 & 0.666 & 0.046 & 0.485  & 0.418 & 0.472 & 0.708 & 0.027\\
  & UA-MT~\cite{yu2019uncertainty} & 0.459 & 0.393 & 0.441 & 0.693 & 0.029 & 0.541 &  0.473  & 0.523 & 0.738 & 0.024\\
  & URPC~\cite{luo2021efficient} & 0.420 & 0.356 & 0.406 & 0.677 & 0.028 & 0.559 & 0.490 & 0.549 & 0.749 & 0.021\\
  & CLCC~\cite{zhao2022cross}& 0.409 & 0.346 & 0.395 & 0.668 & 0.031 & 0.474 & 0.422 & 0.465 & 0.707 & 0.020\\
  & SLC-Net~\cite{liu2022semi} & 0.431 & 0.374 & 0.409 & 0.682 & 0.030 & 0.566 & 0.503 & 0.543 & 0.753 & 0.024\\
  & MC-Net+~\cite{wu2022mutual} & 0.439 & 0.369 & 0.403 & 0.684 & 0.035 & 0.535 & 0.470 & 0.505 & 0.735 & 0.026\\
  & CDMA~\cite{zhong2023semi} & 0.309 & 0.254 & 0.296 & 0.611 & 0.031 & 0.531 & 0.446 & 0.494 & 0.724 & 0.024\\
  & SCP-Net~\cite{zhang2023self} & 0.394 & 0.323 & 0.355 & 0.656 & 0.045 & 0.571 & 0.500 & 0.543 & 0.747 & 0.023\\
  & MCF~\cite{wang2023mcf} & 0.425 & 0.367 & 0.407 & 0.679 & \textbf{0.025} & 0.487 & 0.425 & 0.469 & 0.707 & 0.022\\
  & \textbf{DEC-Seg} (ours) & \textbf{0.592} & \textbf{0.511} & \textbf{0.558} & \textbf{0.758} & \textbf{0.025} & \textbf{0.634} & \textbf{0.564} & \textbf{0.608} & \textbf{0.789} & \textbf{0.019}\\
  \hline
  \end{tabular}\vspace{-0.15cm}
\end{table*}

\begin{table*}[t!]
  \centering
  \small
  \renewcommand{\arraystretch}{1.1}
  \setlength\tabcolsep{6.0pt}
  \caption{Quantitative results on Brain MRI dataset with $10\%$ or $30\%$ labeled data.
  }\label{tab3:BrainMRI}\vspace{-0.25cm}
  \begin{tabular}{r||ccccc|ccccc}
  \hline
  \multicolumn{1}{c||}{\multirow{2}*{\textbf{Methods}}} & \multicolumn{5}{c|}{10\% labeled} & \multicolumn{5}{c}{30\% labeled}\\
  \cline{2-11}
  \multicolumn{1}{c||}{} & mDice $\uparrow$ & mIoU $\uparrow$ &  $F_\beta^w$ $\uparrow$ & $S_{\alpha}$ $\uparrow$ & MAE $\downarrow$ & mDice $\uparrow$ & mIoU $\uparrow$  & $F_\beta^w$ $\uparrow$ & $S_{\alpha}$ $\uparrow$ & MAE $\downarrow$\\
    \hline
   MT~\cite{tarvainen2017mean} & 0.491 & 0.404 & 0.512 & 0.705 & 0.015 & 0.674 & 0.561 & 0.704 & 0.792 & 0.012\\
   DAN~\cite{zhang2017deep} & 0.534 & 0.429 & 0.564 & 0.719 & 0.015 & 0.703 & 0.582 & 0.737 & 0.803 & 0.011\\
   UA-MT~\cite{yu2019uncertainty} & 0.549 & 0.435 & 0.584 & 0.721 & 0.015 & 0.703 & 0.586 & 0.736 & 0.807 & 0.011\\
   URPC~\cite{luo2021efficient} & 0.547 & 0.448 & 0.571 & 0.729 & 0.014 & 0.653 & 0.548 & 0.677 & 0.785 & 0.012\\
   CLCC~\cite{zhao2022cross} & 0.595 & 0.505 & 0.615 & 0.760 & \textbf{0.012} & 0.728 & 0.615 & 0.754 & 0.824 & 0.010\\
   SLC-Net~\cite{liu2022semi} & 0.483 & 0.400 & 0.506 & 0.704 & 0.015 & 0.702 & 0.591 & 0.714 & 0.807 & 0.010\\
   MC-Net+~\cite{wu2022mutual} & 0.580 & 0.481 & 0.595 & 0.749 & 0.015 & 0.746 & 0.640 & 0.766 & 0.834 & 0.009\\
   CDMA~\cite{zhong2023semi} & 0.495 & 0.408 & 0.508 & 0.705 & 0.015 & 0.627 & 0.526 & 0.647 & 0.775 & 0.011\\
   SCP-Net~\cite{zhang2023self} & 0.559 & 0.467 & 0.571 & 0.741 & 0.014 & 0.643 & 0.544 & 0.660 & 0.783 & 0.012\\
   MCF~\cite{wang2023mcf} & 0.589 & 0.491 & 0.623 & 0.750 & 0.013 & 0.704 & 0.612 & 0.722 & 0.819 & 0.010\\
   \textbf{DEC-Seg} (ours) & \textbf{0.626} & \textbf{0.533} & \textbf{0.639} & \textbf{0.777} & \textbf{0.012} & \textbf{0.767} & \textbf{0.671} & \textbf{0.783} & \textbf{0.850} & \textbf{0.008}\\
  \hline
  \end{tabular}\vspace{-0.15cm}
\end{table*}

\begin{table*}[t!]
  \centering
  \small
  \renewcommand{\arraystretch}{1.1}
  \setlength\tabcolsep{6.0pt}
  \caption{Quantitative results on ISIC-2018 dataset with $10\%$ or $30\%$ labeled data.
  }\label{tab4:ISIC2018}\vspace{-0.25cm}
  \begin{tabular}{r||ccccc|ccccc}
  \hline
  \multicolumn{1}{c||}{\multirow{2}*{\textbf{Methods}}} & \multicolumn{5}{c|}{10\% labeled} & \multicolumn{5}{c}{30\% labeled}\\
  \cline{2-11}
  \multicolumn{1}{c||}{} & mDice $\uparrow$ & mIoU $\uparrow$ &  $F_\beta^w$ $\uparrow$ & $S_{\alpha}$ $\uparrow$ & MAE $\downarrow$ & mDice $\uparrow$ & mIoU $\uparrow$  & $F_\beta^w$ $\uparrow$ & $S_{\alpha}$ $\uparrow$ & MAE $\downarrow$\\
    \hline
   MT~\cite{tarvainen2017mean} & 0.822 & 0.738 & 0.809 & 0.811 & 0.104 & 0.844 & 0.760 & 0.822 & 0.825 & 0.087\\
   DAN~\cite{zhang2017deep} & 0.831 & 0.744 & 0.817 & 0.815 & 0.097 & 0.842 & 0.757 & 0.825 & 0.824 & 0.091\\
   UA-MT~\cite{yu2019uncertainty} & 0.839 & 0.752 & 0.818 & 0.820 & 0.094 & 0.845 & 0.762 & 0.829 & 0.829 & 0.086\\
   URPC~\cite{luo2021efficient} & 0.847 & 0.763 & 0.829 & 0.831 & 0.089 & 0.853 & 0.772 & 0.835 & 0.880 & 0.082\\
   CLCC~\cite{zhao2022cross} & 0.842 & 0.756 & 0.815 & 0.820 & 0.088 & 0.845 & 0.762 & 0.816 & 0.821 & 0.092\\
   SLC-Net~\cite{liu2022semi} & 0.843 & 0.754 & 0.818 & 0.822 & 0.091 & 0.846 & 0.752 & 0.811 & 0.821 & 0.085\\
   MC-Net+~\cite{wu2022mutual} & 0.848 & 0.767 & 0.829 & 0.832 & 0.087 & 0.862 & 0.782 & 0.841 & 0.842 & 0.076\\
   CDMA~\cite{zhong2023semi} & 0.858 & 0.777 & 0.842 & 0.839 & 0.083 & 0.868 & 0.788 & 0.848 & 0.846 & 0.073\\
   SCP-Net~\cite{zhang2023self} & 0.852 & 0.770 & 0.835 & 0.835 & 0.085 & 0.853 & 0.769 & 0.831 & 0.834 & 0.083\\
   MCF~\cite{wang2023mcf} & 0.856 & 0.773 & 0.837 & 0.836 & 0.083 & 0.861 & 0.778 & 0.839 & 0.840 & 0.077\\
   \textbf{DEC-Seg} (ours) & \textbf{0.867} & \textbf{0.796} & \textbf{0.858} & \textbf{0.851} & \textbf{0.078} & \textbf{0.875} & \textbf{0.801} & \textbf{0.862} & \textbf{0.856} & \textbf{0.069}\\
  \hline
  \end{tabular}\vspace{-0.15cm}
\end{table*}

\subsection{Comparison with State-of-the-arts}

We compare the proposed DEC-Seg with ten state-of-the-art semi-supervised segmentation methods, \ie,  MT~\cite{tarvainen2017mean}, DAN~\cite{zhang2017deep}, UA-MT~\cite{yu2019uncertainty}, URPC~\cite{luo2021efficient}, CLCC~\cite{zhao2022cross},  SLC-Net~\cite{liu2022semi}, MC-Net+~\cite{wu2022mutual}, CDMA~\cite{zhong2023semi}, SCP-Net~\cite{zhang2023self} and MCF~\cite{wang2023mcf}. For a fair comparison, we change the backbones of all compared methods to ``Res2Net", and then train all compared methods without any data augmentation. All the compared methods undergo $10,000$ iterations, similar to our approach, utilizing the same data and image dimensions. In the case of CLCC, we maintain its original setting of $320 \times 320$ due to the specificity of the method. Moreover, the optimizer, learning rate, and descent strategy all remain consistent with their original configurations across the compared methods.

\subsubsection{Experiments on Polyp Segmentation}

Quantitative comparisons are reported in \tabref{tab1:seen} and \tabref{tab2:unseen}. To validate the learning ability of the two seen training datasets (CVC-ClinicDB and Kvasir), it can be seen from \tabref{tab1:seen} that while most methods learn well, our method performs better than other compared methods, both on $10\%$ and $30\%$ of labeled data, which indicate that our DEC-Seg can fully leverage unlabeled data to improve the polyp segmentation performance. Additionally, we report the comparison results on two unseen datasets (ETIS, and CVC-ColonDB) from \tabref{tab2:unseen} to verify the generalization ability of our DEC-Seg. The CVC-ColonDB and ETIS datasets are difficult, as polyps are with large variations in size. As reported in \tabref{tab2:unseen}, comparing the SCP-Net method on the CVC-ColonDB dataset using $30\%$ labeled data, our method achieves $3.9\%$ and $4.1\%$ improvements on mDice and mIoU, respectively. Similarly, on the ETIS dataset, our performance has a more significant superiority over other methods. The primary reason is that our exploration and utilization of scale information enhance the model's robustness to scale variations. This capability allows it to small polyps, while other methods cannot identify them accurately. Overall, the results indicate that our DEC-Seg has a better generalization ability and effectively utilizes unlabeled data to boost polyp segmentation. 

\begin{figure*}[t]
	\centering
	\begin{overpic}[width=1.0\textwidth]{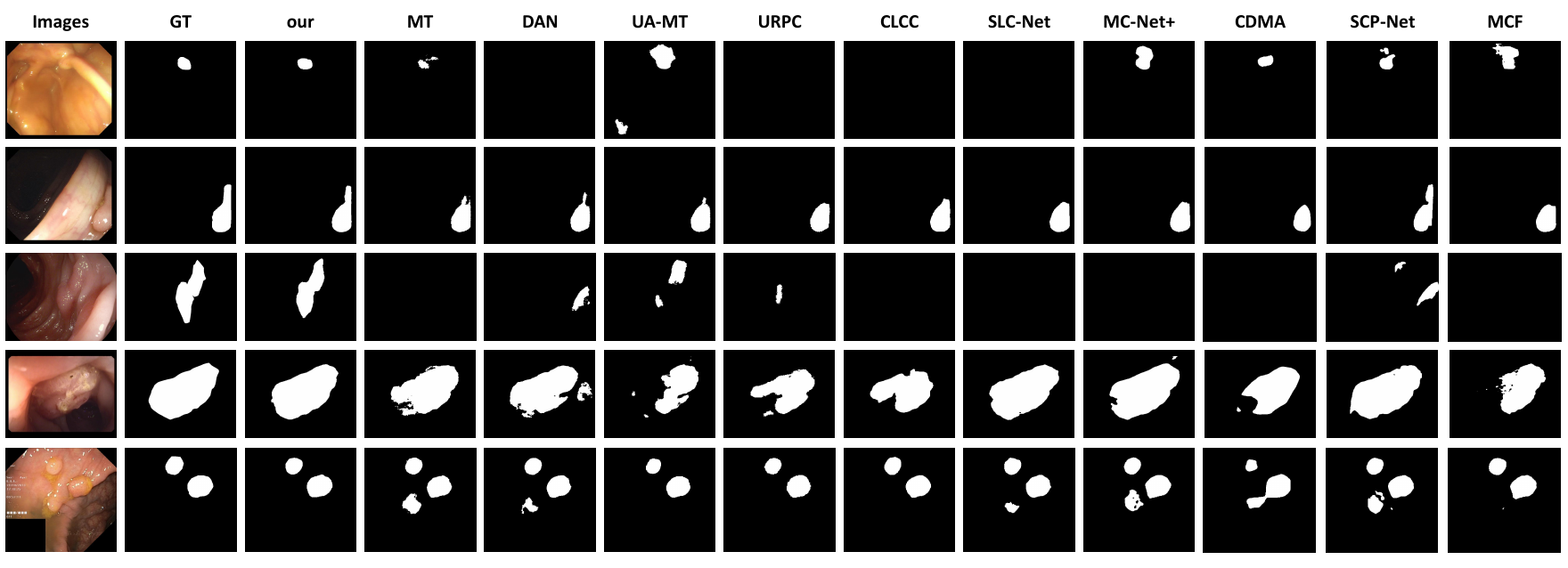}
    \end{overpic}
    \vspace{-0.85cm}
	\caption{{Qualitative results of our model and other ten compared semi-supervised segmentation methods on polyp segmentation datasets.} 
	}
 \label{fig6:polyp_result}\vspace{-0.04cm}
\end{figure*}

\figref{fig6:polyp_result} presents some of the visualization results of the test examples under $30\%$ labeled data. It can be seen that our method can accurately locate and segment polyps under different challenging factors. For example, in the first row of \figref{fig6:polyp_result}, the polyps have very small sizes. It can be seen that DAN, URPC, CLCC, and SLC-Net methods fail to locate polyps. Other methods produce over-segmented fragments and confuse the edges of polyps, while our method accurately and completely segments polyps. In the $3^{rd}$ row, the polyps are visually embedded in their surrounding mucosa, thus it is very difficult to accurately locate and segment these polyps. From the results, our method performs better than other comparison methods to accurately segment polyps. In the $4^{th}$ row, the polyps have relatively large sizes, making it challenging to complete locate the polyps. In this case, some methods (\eg, UA-MT, URPC, CLCC, CDMA, and MCF) only locate some fragments of polyps, while our method can obtain promising segmentation results and produce fine details of the boundary. This is mainly because our method makes use of multi-scale information and cross-generative consistency to learn more powerful feature representations, which help boost the segmentation performance. In addition, we also show some segmentation results containing multiple polyps (see the $5^{th}$ row). 
Compared with other methods, the segmentation of multiple polyps by our method is more complete and accurate. It can be also observed that our method can effectively locate and segment polyps under different challenging factors, such as scale variation, homogeneous regions, non-sharp boundaries, and multiple polyps.

\begin{figure*}[t]
	\centering
	\begin{overpic}[width=1.0\textwidth]{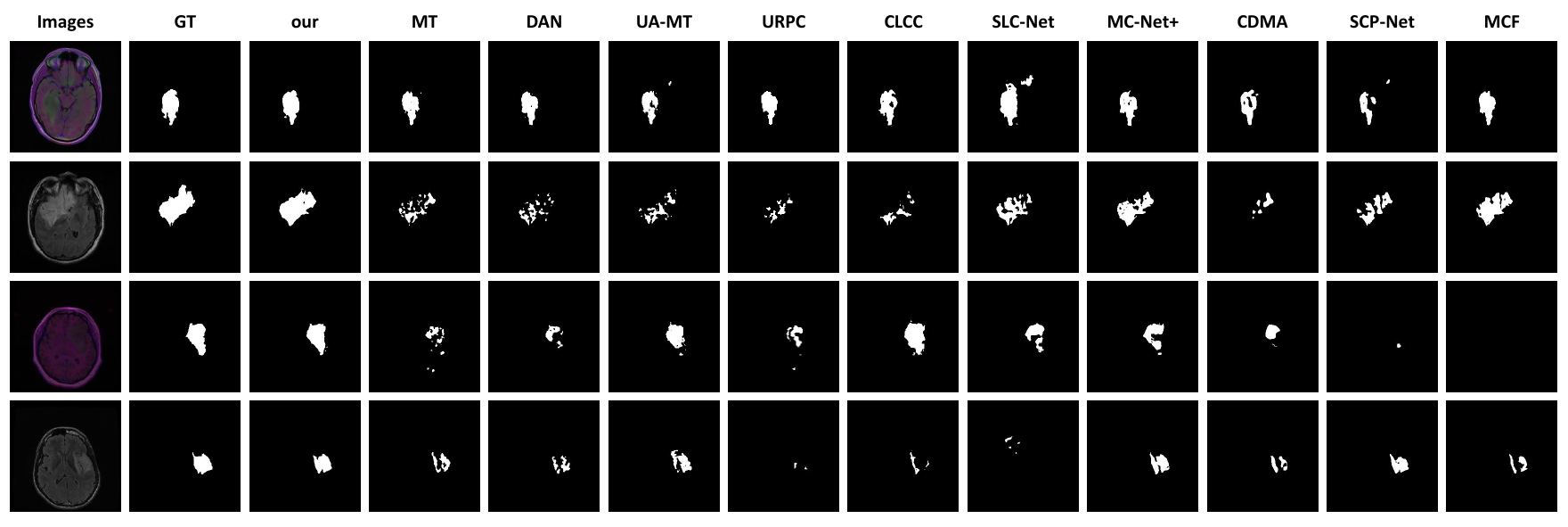}
    \end{overpic}
    \vspace{-0.95cm}
	\caption{{Qualitative results of our model and other ten compared semi-supervised segmentation methods on brain tumor segmentation dataset.} 
	}
    \label{fig7:BrainMRI_result}\vspace{-0.04cm}
\end{figure*}

\begin{figure*}[t]
	\centering
	\begin{overpic}[width=1.0\textwidth]{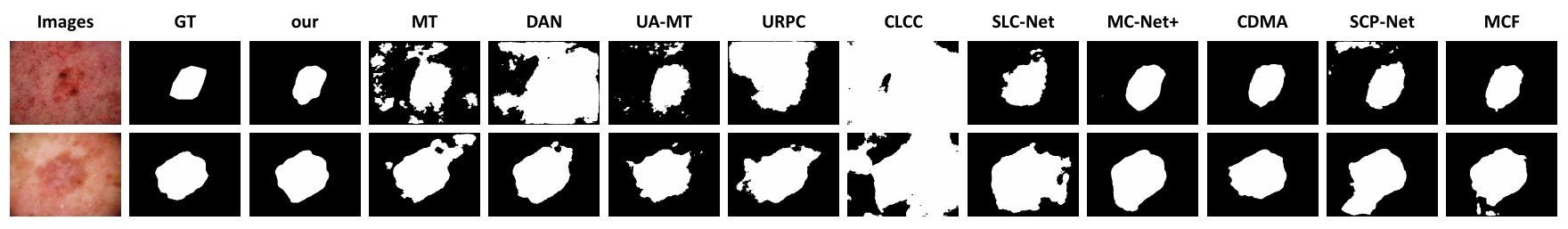}
    \end{overpic}
    \vspace{-0.95cm}
	\caption{{Qualitative results of our model and other ten compared semi-supervised segmentation methods on skin lesion segmentation dataset.} }\label{fig8:ISIC2018_result}\vspace{-0.04cm}
\end{figure*}

\subsubsection{Experiments on Brain Tumor Segmentation}
As shown in \tabref{tab3:BrainMRI}, our DEC-Seg achieves significant improvement in five metrics, and compared to the second best, our method achieves $5.2\%$ and $5.5\%$ improvements in terms of mDice and mIou with $10\%$ labeled data, and the corresponding improvements are $2.8\%$ and $4.8\%$ with $30\%$ labeled data. Moreover, it can be seen in \figref{fig7:BrainMRI_result} that our method can segment the glioma completely and coherently. In the $2^{rd}$ row, the segmentation maps of other methods are discrete, while our method produces more coherent segmented fragments. In addition, in the $4^{th}$ row, some methods (\eg, URPC, CLCC, and SLC-Net) do not find glioma regions, while our method accurately locates and segments glioma regions.

\subsubsection{Experiments on Skin Lesion Segmentation}
\tabref{tab4:ISIC2018} shows the performance on skin lesion segmentation dataset with $10\%$ and $30\%$ labeled ratios. DEC-Seg surpasses all state-of-the-arts. Our method achieves $1.3\%$ and $3.0\%$ improvements over MCF in terms of mDice and mIou with $10\%$ labeled data, and the corresponding improvements are $1.6\%$ and $3.0\%$ with $30\%$ labeled data. Besides, \figref{fig8:ISIC2018_result} shows the visual segmentation results in the skin lesion segmentation dataset. Compared with other methods, our DEC-Seg achieves better and more accurate segmentation in blurred boundaries.

\begin{table*}[!t]
\centering
\small
\renewcommand{\arraystretch}{1.2}
\renewcommand{\tabcolsep}{4.8pt}
\caption{\small{Ablative results on the Kvasir and CVC-ColonDB datasets with $30\%$ labeled data.}
}\vspace{-0.15cm}
\begin{tabular}{l||cccc|cccc}
 \hline
 \multirow{2}*{\textbf{Settings}}  & \multicolumn{4}{c|}{Kvasir} &\multicolumn{4}{c}{CVC-ColonDB}   \\

 \cline{2-9}
 &  mDice  &  mIoU  &  $F_\beta^w$ & $S_{\alpha}$ &  mDice  &  mIoU  &  $F_\beta^w$ & $S_{\alpha}$ \\
 \hline
 (No.1) Baseline & 0.850 & 0.778 & 0.829 & 0.873  & 0.612 & 0.539 & 0.600 & 0.760 \\ 
 (No.2) Baseline + SC & 0.873 & 0.812 & 0.865 & 0.894  & 0.637 & 0.572 & 0.632 & 0.773 \\
 (No.3) Baseline + SC + DCF &  0.882 & 0.818 & 0.874 & 0.892 & 0.675 & 0.599 & 0.658 & 0.790 \\
 (No.4) Baseline + SC + DCF + CC & 0.888 & 0.827 & 0.882 & 0.900 & 0.702 & 0.626 & 0.694 & 0.807 \\
 (No.5) Baseline + SC + DCF + CFA & 0.888 & 0.824 & 0.878 & 0.898 & 0.698 & 0.621 & 0.688 & 0.804 \\
 (No.6) Baseline + SC + DCF + CC + CFA (ours) & \textbf{0.893} & \textbf{0.830} & \textbf{0.886} & \textbf{0.902}  & \textbf{0.721} & \textbf{0.640} & \textbf{0.709} & \textbf{0.814} \\
 \hline
\end{tabular}\label{tab5:abl}
\end{table*}

\subsection{Ablation Study}
To verify the effectiveness of each key component in DEC-Seg, we conduct ablation studies with $30\%$ labeled data on Kvasir and CVC-ColonDB datasets. The ablative results are shown in \tabref{tab5:abl}, where ``Baseline” denotes the semi-supervised framework with only scale-aware perturbation consistency.

\textbf{Effectiveness of SC}. To investigate the importance of scale-enhanced consistency (SC), we add the SC loss to encourage the consistency of predictions from the same inputs with different scales. From \tabref{tab5:abl}, we observe that No.2 (Baseline + SC) outperforms No.1 and obtains a $4.1\%$ improvement in mean Dice. This result indicates that scale-enhanced consistency is very helpful in improving polyp segmentation performance and is robust to scale variation.

\begin{figure}[t!]
	\centering
    \footnotesize
	\begin{overpic}[width=0.95\columnwidth]{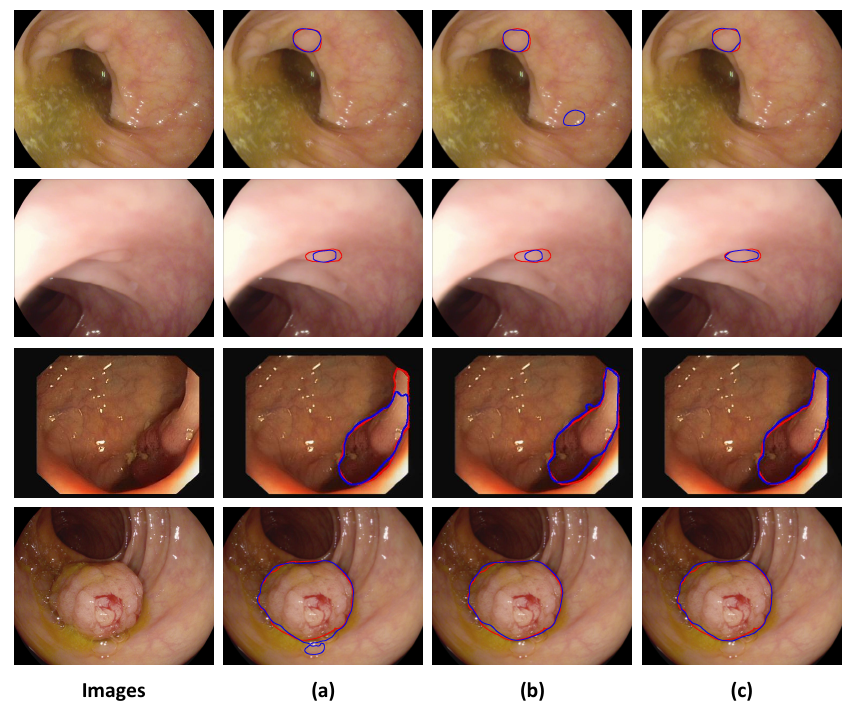}
    \end{overpic}\vspace{-0.15cm}
	\caption{\footnotesize Visualization results of the three decoders. (a) predictions of the decoder (\ie, $D_{1}$) using the original scale features; (b) predictions of the decoder (\ie, $D_{2}$) using the downsampled scale features; (c) predictions of the scale-fused decoders (\ie, $D_f$) using the multi-scale integrated features. The red and blue lines denote the ground truth and predictions, respectively.}\vspace{-0.15cm}
    \label{fig9:three_predication}
\end{figure}

\textbf{Effectiveness of DCF}. As shown in \tabref{tab5:abl}, No.3 (using the proposed DCF module) outperforms No.2 on two datasets. This indicates that the fusion of multi-scale features can further improve the segmentation performance. In addition, to further validate the effectiveness of the dual-scale complement fusion strategy in the proposed DCF module, we construct a ``Basic" strategy, which conducts a concatenation operation followed by two convolution layers to integrate the two features from different scales. The comparison results are shown in \tabref{tab6:abl2}. From \tabref{tab6:abl2}, it can be observed that our DCF performs better than the ``Basic" strategy, indicating the effectiveness of the designed dual-scale complementary fusion module. 
Moreover, we visualize the segmentation results by using three different decoders, \ie, the scale-specific and scale-fused decoders, and the comparison results are shown in \figref{fig9:three_predication}. It can be observed that our method can not accurately locate the boundaries of the polyps when only using the original scale features or the downsampled scale features. However, we integrate the features from the two scales and then propagate them into the scale-fused decoder, which can produce more accurate segmentation maps (as shown in \figref{fig9:three_predication} (c)). This further confirms that integrating features from different scales enhances segmentation performance.

\textbf{Effectiveness of CC}. We further study the contributions of cross-generative consistency (CC). As shown in \tabref{tab5:abl}, it can be seen that No.4 improves the No.3 performance on the CVC-ColonDB dataset, as the mean Dice is improved from $0.675$ to $0.702$. 
Therefore, these improvements indicate that introducing cross-generative consistency loss can help accurately segment polyp tissues in the learning of details and textures. 

\textbf{Effectiveness of CFA}. We then examine the significance of the proposed CFA module. To do this, we integrate the CFA into configurations No.3 and No.4 by aggregating the features of adjacent layers in the encoder before passing them to the decoder. As shown in \tabref{tab5:abl}, No.5 outperforms No.3, and No.6 shows a substantial improvement over No.4 on the CVC-ColonDB dataset, which has significant scale variation. This demonstrates that the CFA module effectively captures scale information and enhances segmentation performance.

\begin{table}[t!]
\centering
\small
\renewcommand{\arraystretch}{1.2}
\renewcommand{\tabcolsep}{6.0pt}
\caption{\small{Ablative results of our DCF module and a basic fusion strategy on the Kvasir and CVC-ColonDB datasets. The labeled ratio is set to $30\%$.}\vspace{-0.15cm}
}
\begin{tabular}{l|ccc|ccc}
 \hline
 \multirow{2}*{\textbf{Settings}}  & \multicolumn{3}{c|}{Kvasir} &\multicolumn{3}{c}{CVC-ColonDB}   \\
 \cline{2-7}
 &  mDice  &  mIoU  &  $F_\beta^w$  &  mDice  &  mIoU  &  $F_\beta^w$ \\
 \hline
 Basic  & 0.885 & 0.823 & 0.878 & 0.710 & 0.633 & 0.703 \\
 DCF & \textbf{0.893} & \textbf{0.830} & \textbf{0.886} & \textbf{0.721} & \textbf{0.640} & \textbf{0.709} \\
 \hline
\end{tabular}\label{tab6:abl2}
\end{table}

\begin{figure}[t!]
	\centering
    \footnotesize
	\begin{overpic}[width=0.85\columnwidth]{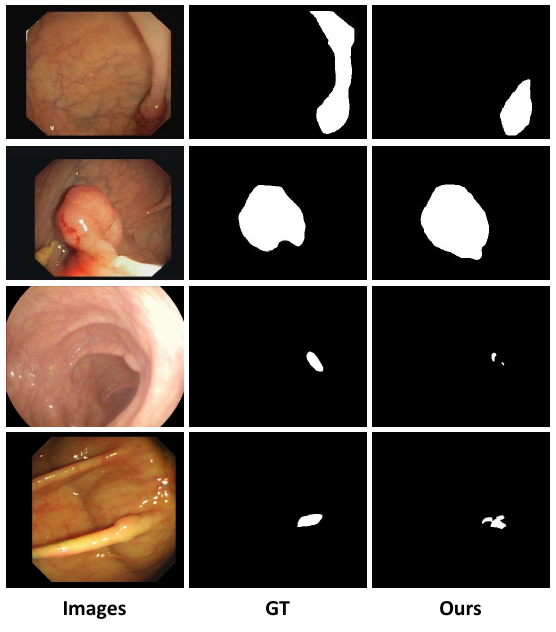}
    \end{overpic}\vspace{-0.15cm}
	\caption{\footnotesize Some failure cases generated by our DEC-Seg.}\vspace{-0.15cm}
    \label{fig10:fail}
\end{figure}

\subsection{Limitation and Future Work}

The qualitative and quantitative evaluations confirm the effectiveness and superiority of our model. However, DEC-Seg encounters challenges in accurately segmenting polyps when dealing with ambiguous areas and unclear boundaries. Some failure cases generated by our model are depicted in \figref{fig10:fail}. In the $1^{st}$ and $2^{nd}$ rows, while the prominent polyp regions are segmented accurately, there are inaccuracies in parts areas of the segmentation. In the $3^{rd}$ and $4^{th}$ rows, it is evident that the polyp regions are notably small and masked within the background, featuring very unclear boundaries, which presents a significant challenge in accurately segmenting the polyp regions. It is apparent that under these circumstances, our DEC-Seg struggles to identify and segment the polyps. Therefore, dealing with ambiguous areas and segmenting polyps with unclear boundaries is desired to be investigated in future work. Additionally, unlike a simple UNet framework, our model may have higher complexity due to incorporating three decoders and two generative networks. Furthermore, our model has only been validated on 2D segmentation tasks. Moving forward, we intend to address and overcome these limitations in our future research endeavors. 

\section{Conclusion}
\label{Conclusion}

In this paper, we have presented a novel semi-supervised learning framework (DEC-Seg) for medical image segmentation. The proposed cross-level feature aggregation module integrates the adjacent features from different resolutions, to enhance the features' representation ability. Then, a scale-enhanced consistency is proposed to handle scale variation and learn more scale-aware features. Meanwhile, we design the scale-fused decoders and a dual-scale complementary fusion module to aggregate the features from the scale-specific decoders and produce the final segmentation maps. Moreover, multiple consistency strategies, \ie, scale-aware perturbation consistency and cross-generative consistency, are presented to enhance the learning process and fully leverage unlabeled data to boost the segmentation performance. Experimental results on multiple datasets from three medical image segmentation tasks show that our DEC-Seg is superior to state-of-the-art semi-supervised segmentation methods.


\ifCLASSOPTIONcaptionsoff
  \newpage
\fi

\bibliographystyle{IEEEtran}
\bibliography{main}

\end{document}